\pdfoutput=1
\documentclass[11pt]{article}

\usepackage{acl}

\usepackage{graphicx}
\usepackage{scalerel} 

\usepackage{hyperref}
\usepackage{url}
\usepackage{arydshln}
\usepackage{color}
\usepackage{colortbl}
\usepackage{booktabs}
\usepackage{adjustbox}
\usepackage{multirow}
\usepackage{wrapfig}
\usepackage{amsmath}

\usepackage{times}
\usepackage{latexsym}

\usepackage[T1]{fontenc}
\usepackage[utf8]{inputenc}

\usepackage{microtype}

\usepackage{inconsolata}
\definecolor{ForestGreen}{RGB}{34,170,34}
\definecolor{CornflowerBlue}{RGB}{90,120,200}
\definecolor{Thistle}{RGB}{245,130,134}

\usepackage{relsize}
\usepackage{array}

\usepackage{pifont}

\newcommand{\guidex}[1]{\textsc{GuideX}\textsubscript{#1}}
\title{\guidex{}: Guided Synthetic Data Generation for Zero-Shot \\Information Extraction} %hau datumultzoan zentratzeko
\author{
\begin{tabular}{cccc}
Neil De La Fuente\textsuperscript{1,3} &
Oscar Sainz\textsuperscript{1,2} &
Iker García-Ferrero\textsuperscript{1,2} &
Eneko Agirre\textsuperscript{1,2} \\
\end{tabular} \\[0.4em] 
\textsuperscript{1}HiTZ Basque Center for Language Technology - Ixa NLP Group \\
\textsuperscript{2}University of the Basque Country (UPV/EHU) \quad
\textsuperscript{3}Technical University of Munich (TUM) \\[0.4em]
\texttt{neil.de@tum.de}
}

\begin{document}
\maketitle
\begin{abstract}

Information Extraction (IE) systems are traditionally domain-specific, requiring costly adaptation that involves expert schema design, data annotation, and model training. While Large Language Models have shown promise in zero-shot IE, performance degrades significantly in unseen domains where label definitions differ. This paper introduces \guidex{}, a novel method that automatically defines domain-specific schemas, infers guidelines, and generates synthetically labeled instances, allowing for better out-of-domain generalization. Fine-tuning Llama 3.1 with \guidex{} sets a new state-of-the-art across seven zero-shot Named Entity Recognition benchmarks. Models trained with \guidex{} gain up to 7 F1 points over previous methods without human-labeled data, and nearly 2 F1 points higher when combined with it. Models trained on \guidex{} demonstrate enhanced comprehension of complex, domain-specific annotation schemas. Code, models, and synthetic datasets are available at \href{https://neilus03.github.io/guidex.com}{neilus03.github.io/guidex.com} 

% We present \guidex{}, a novel framework for synthetic data generation to enhance zero-shot Information Extraction (IE) across diverse and unseen domains. Unlike traditional approaches such as distant supervision and distillation, \guidex{} dynamically generates both annotation guidelines and labeled examples, ensuring schema consistency and reducing annotation noise. This structured guidance enables models to generalize effectively across different schemas without prior exposure. Fine-tuning Llama 3.1 with \guidex{}-generated data leads to an average 10-point F1 improvement across seven zero-shot Named Entity Recognition benchmarks. Furthermore, integrating \guidex{} with existing state-of-the-art methods provides an additional 4-point F1 gain, setting a new state-of-the-art performance. Our error analysis shows substantial improvements in domain adaptability, with models trained on \guidex{} demonstrating enhanced comprehension of complex, domain-specific annotation schemas. We will release the code, models, and synthetic datasets upon acceptance.

\end{abstract}

\section{Introduction}

% Information Extraction (IE) tasks \cite{Grishman1997InformationET} are defined by a formal schema of the entities and relations to be extracted, together with a human-readable guidelines with explanations of what each entity and relation are. IE systems are notorious for the cost to adapt to new domains, which involves domain experts to define the formal schema and the detailed guidelines, a team of annotators to annotate text according to the schema and guidelines, and machine learning experts to train a system

% %highly dependent on specific domains, typically guided by closed schemas and detailed guidelines. Systems developed for these tasks have been schema-dependent and often unable to function outside their designated training label space. 

Information Extraction (IE) tasks \cite{Grishman1997InformationET} are structured around two core components: a formal schema specifying target entities/relations and human-readable guidelines defining their interpretation. Despite their utility, IE systems face significant scalability challenges due to domain dependence. Adapting to new domains requires substantial resources, including (1) domain experts to design schemas and annotation rules, (2) trained annotators to label data accordingly, and (3) machine learning specialists to develop performant models. This complex process creates a bottleneck for real-world applications where label definitions frequently evolve across contexts.

% Early attempts at removing the need for annotated data (zero-shot IE) approached the task as either Question Answering~\cite{levy-etal-2017-zero} or Natural Language Inference~\cite{obamuyide-vlachos-2018-zero, sainz2021label}, utilizing additional supervised data from these related tasks. The results were promising, but they still required carefully constructed manual schemas and guidelines, and failed to generalise to new domains.

Early attempts to remove the need for annotated data (zero-shot IE) framed the task through Question Answering~\cite{levy-etal-2017-zero} or Natural Language Inference~\cite{obamuyide-vlachos-2018-zero, sainz2021label} paradigms, leveraging supervised data from these auxiliary tasks. While showing initial promise, these methods remained constrained by their reliance on manually crafted schemas and limited cross-domain generalization. 

% %
% More recent approaches leverage Large Language Models (LLMs) to reduce the cost of defining the schema and guidelines, with improved performance ~\cite{sainz2024gollie, li2024knowcoder}, making it easier to move to new domains and schemas.  However, these systems still underperform when applied to an unseen domain where the definition of the known labels might change~\cite{sainz2024gollie}. In this work, we aim to bridge the gap between seen and unseen domains by automatically generating synthetic schemas, guidelines and annotated data for the relevant domains of interest.

More recent advancements leveraging Large Language Models (LLMs) have streamlined schema definition processes~\cite{li2024knowcoder}, facilitating adaptation to new domains. However, performance gaps persist when deploying these systems in unseen domains where label semantics diverge from training distributions~\cite{sainz2024gollie}. Our work addresses this challenge by automating the generation of domain-specific schemas, guidelines, and annotated data to bridge the seen-unseen domain divide.

\begin{figure}[!t]
\centering
    \includegraphics[width=1\linewidth]{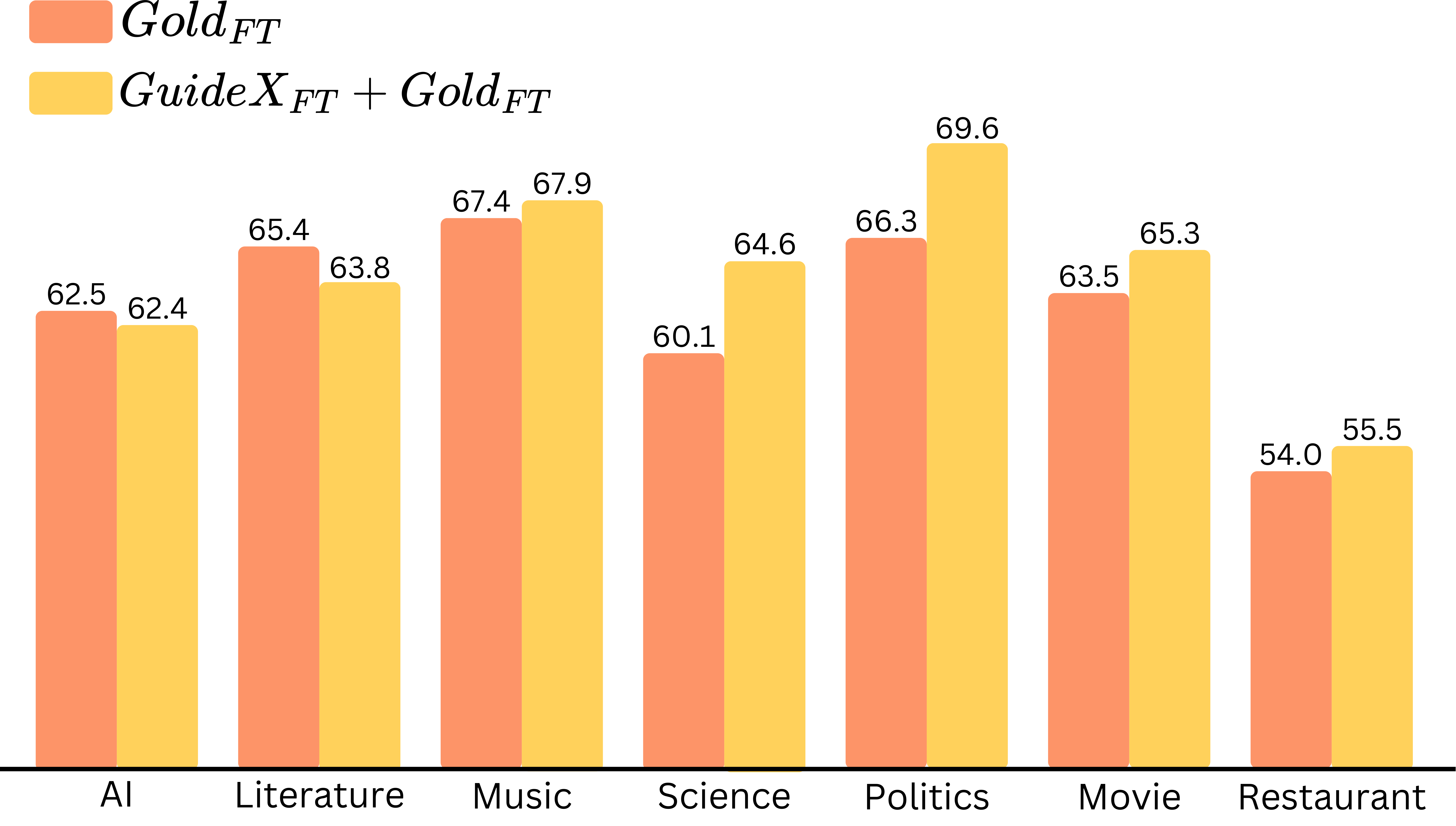}
    \caption{Impact of \guidex{} for zero-shot NER on different domains. In $Gold_{FT}$, the model is trained solely on gold training data, whereas in $GuideX_{FT}+Gold_{FT}$ it is also trained on our synthetic dataset.}
    \label{fig:teaser}
\end{figure}

% Data Augmentation \cite{feng2021survey} and, particularly, Synthetic Data Generation \cite{OpenHermes-2.5,xu2025magpie} have shown to be very useful in the era of LLMs. In the field of IE, techniques such as distant supervision~\cite{mintz2009distant} have been traditionally applied to improve the performance of the models. Unfortunately, most of these techniques introduce noisy annotations. Distant supervision, while effective in achieving high recall, tends to generate a substantial amount of spurious annotations. LLM distillation \cite{hinton2015distilling}, where a smaller more efficient student model learns from the predictions of a bigger teacher model, instead, is usually upper-bounded by the teacher's performance, and fails to completely annotate all the correct instances. Furthermore, both methods require a predefined annotation schema, limiting the practical applicability to unseen domains.

Data Augmentation \cite{feng2021survey} and Synthetic Data Generation \cite{OpenHermes-2.5,xu2025magpie} have proven particularly valuable in the era of LLMs. In IE, traditional techniques like distant supervision~\cite{mintz2009distant} aim to enhance model performance but often introduce noisy annotations. While effective for achieving high recall, distant supervision frequently generates spurious labels due to its reliance on imperfect heuristics. Similarly, LLM distillation~\cite{hinton2015distilling}—where smaller student models learn from larger teacher models—faces inherent limitations: student performance is constrained by the teacher's capabilities and often fails to capture all valid instances. Both methods also require predefined annotation schemas, severely limiting their adaptability to novel domains where label definitions may shift.

In this paper, we tackle the aforementioned limitations by introducing \guidex{}, a novel data generation method inspired by the work of domain experts. It is  designed to \textbf{generate schemas, guidelines and annotated examples for any new domain} which allows to improve IE performance on new as well as on unseen domains. %reduce annotation noise while maintaining a diverse annotation schema}. 
This approach consists of four main steps. Given a set of documents from the target domain, an LLM is used to identify the key information within each document, summarizing and synthesizing its content into a set of bulleted ideas. Next, the extracted information is structured into a standardized format, typically a JSON file. Then, the model is asked to generate the annotation schema and the corresponding annotation guidelines based on the previously structured annotations. This approach ensures that the annotations align with the guidelines and remain comprehensive. This step is particularly crucial, as it guarantees the correctness of the schema and significantly reduces potential annotation errors in the generated data. Finally, we ask the model to generate the final annotations following a standard code-style format.

We validated our data generation approach by training state-of-the-art models using the synthetic data produced through our methodology. When fine-tuning base models, such as Llama 3.1~\cite{grattafiori2024Llama3herdmodels}, exclusively on our dataset, we observed an average improvement of 10 F1 points across seven Named Entity Recognition (NER) benchmarks in zero-shot evaluation. Furthermore, when leveraging our data to enhance state-of-the-art approaches, we achieved a notable improvement of nearly 2 F1 points on the same benchmarks (see Figure~\ref{fig:teaser}), establishing a new state-of-the-art.

% The key contributions of this work are as follows:
% \begin{itemize}
%     \item We propose and validate \guidex{}, a novel synthetic data generation approach tailored for IE.
%     \item Leveraging \guidex{}, we train a state-of-the-art zero-shot NER model, surpassing the previous SoTA by 4.1 F1 points on standard evaluation benchmarks.
%     \item We conduct a comprehensive analysis to evaluate the impact of our synthetic data on guideline adherence, demonstrating its effectiveness in improving model alignment with annotation standards.
% \end{itemize}

% All research artifacts, including code, models, and datasets, will be released upon acceptance to support further research in guideline-aware IE.

% \begin{figure*}[!ht]
%     \centering
%     \includegraphics[width=1\linewidth]{figs/synthetic-pipeline.pdf}
%     \caption{An overview of the \guidex{} pipeline. Given a real-world text (a news article, wikipedia style document, etc) the LLM follows the 4 steps above to generate a pair of annotation guidelines and annotations.}
%     \label{fig:synthetic-data-pipeline}
% \end{figure*}

\section{Related Work}
In this section we review the literature related to our work. We begin by highlighting the most significant studies that involve LLMs for IE. Following that, we will focus on methods for generating synthetic data aimed at the same task.
% The field of IE (IE) has seen remarkable progress with the rise of LLMs (LLMs). However, significant challenges remain in achieving accurate and robust performance, particularly in scenarios involving intricate annotation guidelines and complex real-world data distributions.
% \paragraph{Traditional IE.}
% Early IE approaches relied on either rule-based or supervised models that required extensive handcrafted features and large-scale annotated data \cite{grishman1996message, tjong-kim-sang-de-meulder-2003-introduction}. Distant Supervision \cite{mintz2009distant}, in particular, linked mentions in unannotated text to facts in knowledge bases, automatically labeling them for large-scale training. While this method offered a practical way to scale data labeling, it introduced considerable noise as entity co-occurrence does not necessarily reflect the intended relation or event. Consequently, distant supervision pipelines risk spurious annotations and have trouble adapting to new domains because their label schemas are fixed in advance. Although techniques like multi-instance learning or noise-reduction heuristics have been proposed \cite{surdeanu2012multi}, these only partially mitigate the label mismatch problem.
\paragraph{LLM-Based IE.}
Recent advances in IE increasingly leverage LLMs to tackle tasks such as Named Entity Recognition (NER), Relation Extraction (RE), and Event Extraction (EE) in both zero-shot and few-shot settings~\cite{NEURIPS2020_1457c0d6,10.5555/3455716.3455856,xu2024large}. By prompting \cite{li-etal-2022-prompt-based-text, ashok2023promptnerpromptingnamedentity, wang-etal-2021-zero, wei2024chatiezeroshotinformationextraction, wei2023zero, xu-etal-2024-chatuie, mo-etal-2024-c} or fine-tuning \cite{zhou2023universalner, lou2023universal, 10.1007/978-3-031-77847-6_4} these models can be guided to extract relevant spans (e.g., entities, relationships) in raw text. Methods like InstructUIE~\cite{wang2023instructuiemultitaskinstructiontuning}, and RUIE~\cite{liao-etal-2025-ruie} build on the idea of formulating IE as an instruction-following or retrieval-based generation task, showing that well-structured prompts can improve performance without extensive human-annotated data.

A specialized subtrend emphasizes schema and guideline guided strategies, where the LLM is trained to follow explicit annotation rules \cite{sainz2024gollie, pang-etal-2023-guideline, bai-etal-2024-schema}. These methods demonstrate that providing precise definitions and examples of valid or invalid annotations can significantly enhance zero-shot IE outcomes, reducing error rates from ambiguous token boundaries or unclear entity types. Similarly, KnowCoder~\cite{li2024knowcoder} encodes structured knowledge into LLMs to facilitate universal IE across multiple domains, underscoring the value of carefully specified label schemas. Although these guideline-oriented approaches maintain higher consistency across domains, they often depend on time-consuming, manual curation of instructions. As new tasks or domains emerge, the annotation guidelines must be updated or expanded, posing a key scalability challenge.

\begin{figure*}[!ht]
    \centering
    \includegraphics[width=0.9\linewidth]{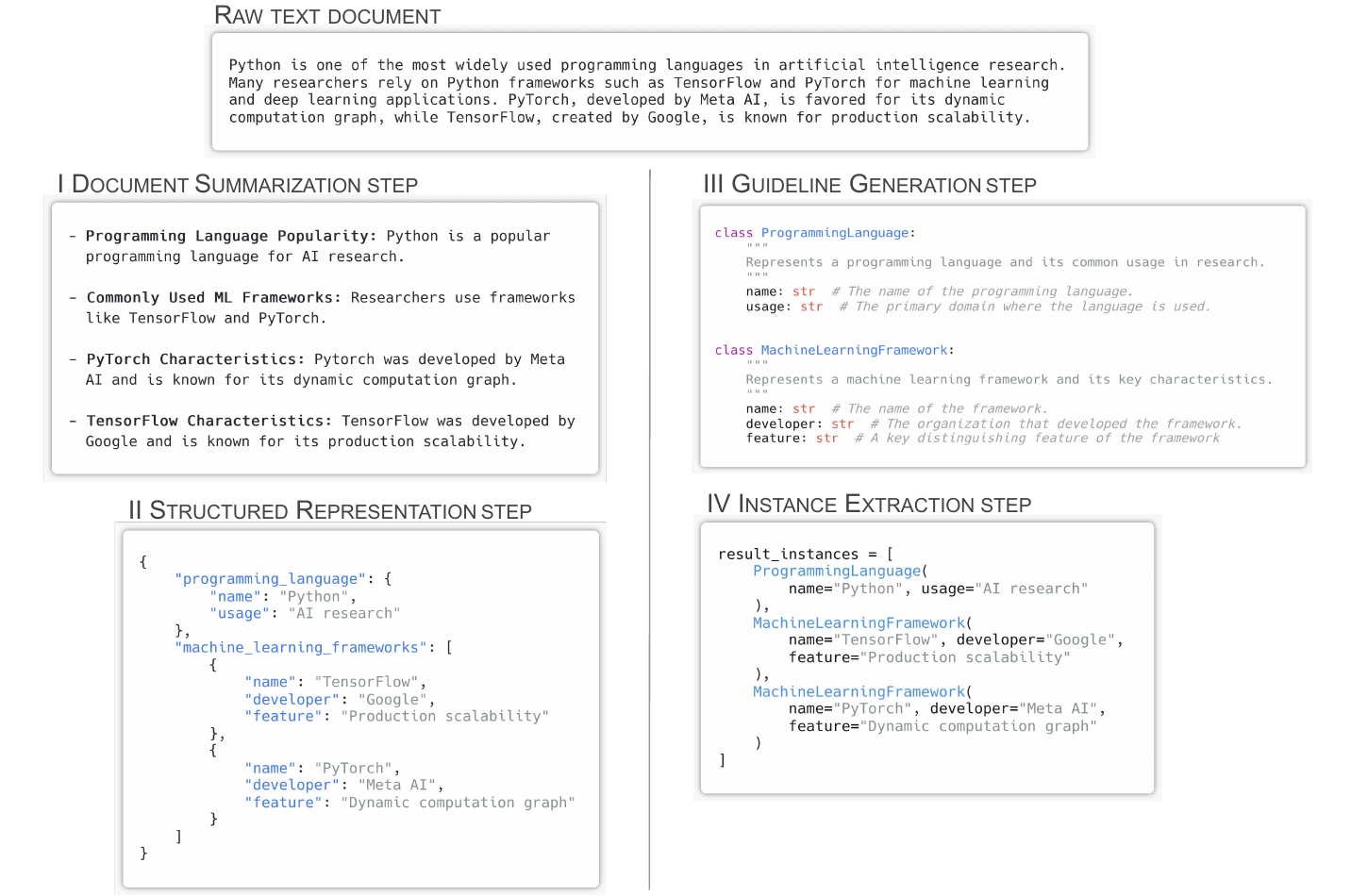}
    \caption{\guidex{} process overview. The approach transforms raw text into structured annotations by dynamically inferring schemas, generating executable guidelines, and resulting annotations.}% The process enables automated validation to reduce annotation noise and ensure consistency.}
    \label{fig:synthetic-data-pipeline}
\end{figure*}

\paragraph{Synthetic Data Generation for IE.} Distant supervision remains one of the earliest forms of synthetic data labeling, aligning knowledge-base facts with textual mentions to automate the creation of large-scale IE datasets~\cite{mintz2009distant}. While this technique helped address the scarcity of annotated data, it is susceptible to label noise because the mere co-occurrence of entities in a sentence does not necessarily confirm their relationship~\cite{surdeanu2012multi}. More recent developments incorporate multi-instance multi-label learning and noise reduction heuristics to mitigate erroneous matches~\cite{hoffmann-etal-2011-knowledge, lin-etal-2016-neural,han2016global, xiao-etal-2020-denoising}, yet these pipelines often rely on rigid predefined schemas, making it challenging to adapt to novel entity or relation types.

Beyond distant supervision, emerging methods leverage LLMs themselves to generate synthetic data for IE tasks~\cite{josifoski-etal-2023-exploiting, chen-etal-2017-automatically}. UniNER~\cite{zhou2023universalner} and KnowCoder~\cite{li2024knowcoder} generate additional training instances by prompting an LLM to produce sentences conforming to a particular label schema. This broadens coverage and reduces the requirement for human annotation. However, the resulting data can suffer from stylistic repetition, domain mismatch, or incomplete label application if the model fails to follow the schema consistently~\cite{xu2024large}. Moreover, existing synthetic data generation techniques often do not provide robust mechanisms to manage data structure or coverage, resulting in partially accurate guidelines and noisy annotations. Complementary lines of work employ information‑theoretic criteria such as \emph{V‑information} to automatically select or weight synthetic examples, demonstrating measurable gains for classification and slot‑filling tasks~\cite{pmlr-v162-ethayarajh22a}. Orthogonally, weak‑supervision pipelines built on data‑programming have been revisited in the LLM era; for instance, \textsc{PromptRE} combines prompting outputs with Snorkel‑style label aggregation to create document‑level relation corpora without manual annotation~\cite{gao-etal-2024-promptre}. These approaches reduce labeling cost but still assume a predefined schema, and thus remain complementary to our automatic schema‑induction strategy.

In light of these observations, our work introduces \guidex{}, a data generation approach that integrates both schema-driven and synthetic paradigms while addressing key limitations in diversity and noise control. Rather than relying on fully manual guidelines or naive LLM-based generation, \guidex{} dynamically constructs annotation schemas and guidelines for each document, then synthesizes labeled text aligned with those rules. This approach reduces the costs of manual schema creation and mitigates the spurious annotations often seen in unconstrained synthetic data generation. Crucially, we show that incorporating the resulting dataset into zero-shot IE training significantly boosts performance across multiple benchmarks, surpassing existing approaches whether they rely on synthetic data and guidelines or not. By unifying explicit guideline creation with data synthesis, \guidex{} offers a %scalable, 
low-noise strategy for robust zero-shot IE across increasingly diverse domains.

\section{\guidex{}: Guided Synthetic Data Generation}

In this section, we introduce \guidex{}, a structured synthetic data generation approach designed to enhance IE capabilities of LLMs. Unlike traditional LLM distillation approaches or distant supervision, which often rely on predefined annotation schemas, \guidex{} dynamically infers annotation schemas. This approach reduces annotation inconsistencies, enhances flexibility across different IE tasks, and enables high-quality, guideline-driven annotations.

As shown in Figure~\ref{fig:synthetic-data-pipeline}, the \guidex{} approach consists of four sequential steps: document summarization, structured representation synthesis, annotation guideline generation, and instance extraction.  Prior work has shown that LLMs struggle with information extraction in zero-shot settings, and attempting to perform the entire task in a single step leads to poor results. Instead, our structured process progressively refines raw text through these four stages, ensuring high-quality, structured annotations that align with inferred guidelines.
Complete prompt templates for the four stages are listed in Figure~\ref{fig:synthetic-data-prompts} in the appendix so that the readers can reproduce every step. Below, we describe each step in detail.

\paragraph{Document summarization.}

The first step in the pipeline focuses on identifying the most important concepts within a document. To achieve this, the LLM generates a summary highlighting the key points, effectively recognizing relevant entities and events while structuring the extracted information. Instead of relying on a predefined annotation schema, this approach allows the model to determine the relevant elements autonomously, resulting in more diverse and domain-specific annotations.

Figure~\ref{fig:synthetic-data-pipeline}.I presents a sample summary generated for an article discussing Machine Learning frameworks in AI research. The model successfully identifies key frameworks (\textit{TensorFlow} and \textit{PyTorch}), along with relevant entity details, including their developers and notable features.

\paragraph{Structured representation.}
IE aims to identify specific spans of text that contain relevant information. In the second step of our approach, the model leverages both the previously generated summary and the original document to organize the extracted information into a structured JSON format. This representation ensures that key elements are systematically categorized while maintaining direct references to the source text. To improve accuracy and conciseness, we enforce constraints that limit the extracted spans to the shortest possible length while preserving their full meaning.

As depicted in Figure~\ref{fig:synthetic-data-pipeline}.II, the extracted information is transformed into a structured JSON format, where each entity is assigned meaningful labels and attributes. This structured representation enables better organization and downstream usability, ensuring that information is both interpretable and machine-readable.

\paragraph{Guideline generation.}

Annotation guidelines play a critical role in improving zero-shot IE performance by ensuring consistency and reducing ambiguity in annotations~\cite{sainz2024gollie, li2024knowcoder}. However, manually crafting such guidelines is a complex task, as it requires defining precise rules that account for variations across different domains. To address this challenge, our approach dynamically generates annotation guidelines based on the structured JSON representation, the document summary, and the original text. Instead of relying on predefined schemas, the model autonomously derives comprehensive descriptions for each entity type, ensuring that all relevant attributes are clearly captured.

Each generated guideline is formulated as a Python dataclass, embedding a long and detailed description of the entity type within the class docstring. The expected attributes of the entity are also explicitly defined, with each field accompanied by comments explaining its meaning and expected format. By structuring the annotation schema in this manner, the model produces a standardized and interpretable representation of the information.

Figure~\ref{fig:synthetic-data-pipeline}.III shows how this process results in a set of structured dataclasses that define both entity types and their relationships while preserving flexibility. Unlike static annotation guidelines, which may be limited in adaptability, this approach ensures that each document receives tailored guidelines that align precisely with its content. By encoding these guidelines into Python code, the approach maintains a high level of structural consistency.

\paragraph{Instance Extraction}

The final step of the \guidex{} approach involves extracting concrete instances of the entities and attributes defined in the annotation guidelines. Using the structured JSON representation and the inferred annotation schema, the model populates entity classes with specific values derived directly from the original document. This step ensures that the extracted instances strictly adhere to the predefined structure and maintain high fidelity to the source text.

To achieve this, the model generates a Python list where each entry corresponds to an instance of one of the dataclasses created in the guideline generation step. Each instance is populated with concise values extracted from the text, prioritizing single words or short phrases over verbose descriptions to maximize precision and clarity. The model is explicitly instructed to return only the structured instances, without additional explanations or extraneous text.

As illustrated in Figure~\ref{fig:synthetic-data-pipeline}.IV, this process results in a structured dataset where each entity and its attributes are instantiated according to the inferred schema. This format ensures seamless integration into downstream applications, enabling models to leverage high-quality, structured annotations for training or evaluation. By enforcing a standardized representation of extracted instances, \guidex{} maintains consistency across different documents and domains

% \paragraph{} 
\guidex{} enhances synthetic data reliability by structuring annotation guidelines and extracted instances as executable Python code, enabling automated validation to detect hallucinations and inconsistencies. A consistency-checking mechanism systematically executes each dataset entry, flagging logical errors such as undefined entity types or misaligned attributes. An automated filtering script discards invalid annotations while retaining only schema-compliant ones, significantly reducing spurious relationships and annotation noise. This structured validation process ensures high-quality, guideline-aligned synthetic data, leading to more robust and trustworthy IE models.

% \paragraph{(Optional) Output reformatting.}

% % \begin{figure}
% %     \centering
% %     \includegraphics[width=.8\linewidth]{figs/carbon_oscar/carbon_annotations.png}
% %     \caption{Generated guidelines and annotations following the python dataclass format.}
% %     \label{fig:python_representation}
% % \end{figure}

% At this stage of the pipeline, the annotations and corresponding guidelines have been generated. However, their structure may differ from conventional IE datasets, which typically follow standardized formats expected by most supervised systems. To ensure compatibility, we adopt the format proposed by \citet{sainz2024gollie}, representing both guidelines and annotations using Python classes and instances. This structured representation facilitates seamless integration with existing models and training pipelines, maintaining consistency across annotation schemes.

% Figure~\ref{fig:python_representation} shows the Python representation of the previous annotations and guidelines. Some of the annotations are independent such as \textsc{Person} or \textsc{Location} while others are directly integrated into a general annotation template \textsc{Detection}. In this step the annotations and labels have change slightly, however, they maintain coherent.

\section{The \guidex{} dataset}

This section presents the dataset constructed using the proposed method. We begin by detailing the document collection process that served as the foundation for our dataset. Then, we provide statistical insights to illustrate its composition and characteristics.

\paragraph{Document Collection.} The dataset was constructed using FineWeb-edu~\cite{penedo2024the}, a high-quality subset of the larger FineWeb dataset, specifically curated for educational content. From this collection, we randomly sampled \textasciitilde 10,000 documents. % based on their educational score. 
The dataset exhibits a wide range of document lengths, spanning from 194 to 22.6k words. To preserve the coherence of the textual structure and maximize contextual understanding, we retain entire documents rather than segmenting them into smaller units such as paragraphs or sentences.

% \begin{table}[]
%     \centering
%     \resizebox{.5\textwidth}{!}{%
%     \begin{tabular}{rl|rl}
%         \toprule
%         \multicolumn{4}{c}{Label distribution} \\
%         \midrule
%         \multicolumn{2}{c|}{Most common} & \multicolumn{2}{c}{Least common} \\
%         Freq. & Label & Freq. & Label \\
%         \midrule
%         1820 & Symptom   & 1 & MusicOrigin \\
%         1459 & Benefit   & 1 & AttitudesTowardsMusic \\
%         929  & Resource  & 1 & MusicStudy \\
%         927  & Topic     & 1 & DietRecommendations \\
%         837  & Cause     & 1 & FilterInformation \\
%         830  & Location  & 1 & AsyncDataSharing \\
%         786  & Event     & 1 & SoundMakingInformation \\
%         689  & Study     & 1 & MOOCDefinition \\
%         679  & Treatment & 1 & MOOCDesing \\
%         609  & HistoricalEvent & 1 & MOOCContent \\
%         586  & Application & 1 & TeachingAndLearning \\
%         572  & Activity  & 1 & BenefitsAndChallenges \\
%         565  & Author    & 1 & AfricanAmericanPortrayal \\
%         \bottomrule
%     \end{tabular}
%     }
%     \caption{Most and least common labels existing in the \guidex{} dataset.}
%     \label{tab:label_statistics}
% \end{table}

\begin{table}[t]
    \centering
    \small                 
    \setlength{\tabcolsep}{6pt}  
    \renewcommand{\arraystretch}{1.1}
    \begin{adjustbox}{width=\columnwidth}
    %\rowcolors{2}{yellow!6}{white}
    \begin{tabular}{r l | r l}
    \toprule
    \multicolumn{2}{c|}{\textbf{Most common}} & \multicolumn{2}{c}{\textbf{Least common}} \\
    \midrule
    \textbf{Freq.} & \textbf{Label} & \textbf{Freq.} & \textbf{Label} \\
    \midrule
    1820 & Symptom             & 1 & MusicOrigin \\
    1459 & Benefit             & 1 & AttitudesTowardsMusic \\
    929  & Resource            & 1 & MusicStudy \\
    927  & Topic               & 1 & DietRecommendations \\
    837  & Cause               & 1 & FilterInformation \\
    830  & Location            & 1 & AsyncDataSharing \\
    786  & Event               & 1 & SoundMakingInformation \\
    689  & Study               & 1 & MOOCDefinition \\
    679  & Treatment           & 1 & MOOCDesign \\
    609  & HistoricalEvent     & 1 & MOOCContent \\
    586  & Application         & 1 & TeachingAndLearning \\
    572  & Activity            & 1 & BenefitsAndChallenges \\
    % 565  & Author              & 1 & AfricanAmericanPortrayal \\
    \bottomrule
    \end{tabular}
    \end{adjustbox}
    \caption{Most and Least frequent labels in the \guidex{} dataset.}
    \label{tab:label_statistics}
\end{table}

\paragraph{Dataset Statistics.} The \guidex{} dataset covers a diverse range of topics, as illustrated by the distribution of the most and least frequent labels in Table~\ref{tab:label_statistics}. It includes a strong presence of categories related to Medicine (\textit{Symptom}, \textit{Cause}, \textit{Study}, \textit{Treatment}), Economics (\textit{Benefit}, \textit{Resource}, \textit{Application}, \textit{Activity}), and History (\textit{Event}, \textit{HistoricalEvent}, \textit{Study}). Additionally, it covers domains such as Music (\textit{MusicOrigin}, \textit{AttitudesTowardsMusic}, \textit{MusicStudy}) and Education (\textit{MOOCDefinition}, \textit{MOOCDesign}, \textit{TeachingAndLearning}). In total, the dataset has 28,677 unique labels, with an average of 5.34 distinct labels per document. Each document contains an average of 11.39 annotations, highlighting the dataset's richness and granularity. % Full dataset statistics, including train/dev/test splits and an entity type overlap analysis between the gold datasets and \guidex{}, are provided in Appendix~\ref{sec:appendix_data_overlap}.

\paragraph{Entity–type overlap with existing corpora.}
We compared the 28,677 unique entity--type names that appear in \guidex{} against the label spaces of 35 widely–used IE datasets covering NER, RE, EE, EAE and SF. Across all \textit{train} splits we found 243 distinct gold labels, of which 103 (\textbf{42.4\%}) are already present verbatim in \guidex{}. A very similar ratio holds for \textit{test} splits (98 / 235 labels, \textbf{41.7\%}). In other words, \guidex{} captures roughly two–fifths of the known label space \emph{without} having been designed for any of those benchmarks.

Coverage is naturally uneven. Generic NER datasets such as \textsc{CoNLL}\,03, BroadTwitter, HarveyNER and BC5CDR achieve \textbf{100\%} type overlap, while the CrossNER family is above 90\%. Conversely, event–centric resources like \textsc{ACE05‐RE}, \textsc{CASIE} and \textsc{E3C} contribute long-tail, highly specialised labels that \guidex{} does not yet include (overlap $\le 15\%$). Taken together, these numbers show that \guidex{} already offers substantial coverage of standard NER schemas while still leaving room for complementary, task-specific labels provided by human annotation. They also explain why models benefitted most on domains whose gold labels are either absent or only partially represented in \guidex{}.

\section{Experimental Setting}

In this section we outline the details of our experiments. We begin by introducing the models used for both synthetic dataset generation and fine-tuning on manually annotated (gold) data. Next, we describe the IE datasets utilized for training and evaluation. Finally, we present the baselines and state-of-the-art systems used for comparison.

\subsection{Models}

\paragraph{Synthetic data generation.} For generating synthetic data, we utilized the 70B variant of the Llama 3.1 Instruct model. After evaluating various alternatives, this model demonstrated consistent reliability in generating high-quality outputs. Although proprietary models exhibited marginally better performance in some cases, we prioritized reproducibility by selecting an open-source solution for this research.

\paragraph{Model fine-tuning.} All the models we trained are based on the 8B variant of Llama 3.1~\cite{grattafiori2024Llama3herdmodels}. We adopted the standard code-style format for IE ~\cite{sainz2024gollie,li2024knowcoder,li2023codeie, qi2024adelie}. To avoid negative impacts caused by the input format discrepancies, we use the base Llama 3.1 over the instruct variant.

\subsection{Datasets}
\label{sec:datasets}
\paragraph{Training Datasets.} Beyond the \guidex{} dataset, we trained our system using the same manually annotated gold-standard data as~\citet{sainz2024gollie}. This gold-standard data comes from multiple sources and covers several IE tasks including Named Entity Recognition (NER), Event Extraction (EE), Event Argument Extraction (EAE), Relation Extraction (RE), and Slot Filling (SF). Specifically, we utilized ACE 2005~\cite{walker2006ace} for NER, EE, EAE, and RE, while TACRED~\cite{zhang2017position} was employed for SF\footnote{Originally designed for RE, we followed \citet{sainz2024gollie} in converting it into an SF task.}. Additional NER datasets used for training include BC5CDR~\cite{li2016biocreative}, CoNLL 2003~\cite{tjong-kim-sang-de-meulder-2003-introduction}, DIANN~\cite{fabregat2018overview}, NCBDisease~\cite{dougan2014ncbi}, Ontonotes 5~\cite{pradhan2013towards}, and WNUT17~\cite{derczynski-etal-2017-results}.

\paragraph{Evaluation Datasets.} To assess the effectiveness of our approach, we conducted evaluations on standard zero-shot NER benchmarks. Specifically, we tested on multiple CrossNER~\cite{liu2020crossner} splits, as well as the MIT Movie and MIT Restaurant datasets~\cite{DBLP:conf/icassp/LiuPCG13}.

\subsection{Baselines} \label{sec:baselines}

% We defined two main baselines for our experiments: Llama 3.1 and Llama 3.1 fine-tuned with the previously mentioned training datasets. From now on, we will refer to the fine-tuned baseline as Gold\textsubscript{FT}. We will compare this two baselines to their equivalents trained first on the \guidex{} dataset, namely \guidex{FT}, and \guidex{FT} + Gold\textsubscript{FT}.

We established two primary baselines: the base Llama 3.1 model and its fine-tuned counterpart using the manually annotated training datasets described in Section~\ref{sec:datasets}. From now on, we refer to the fine-tuned version as Gold\textsubscript{FT}. These baselines will be compared to their variants pretrained on \guidex{} synthetic data: \guidex{FT} (trained solely on synthetic data) and \guidex{FT} + Gold\textsubscript{FT} (trained sequentially on both datasets).

%Our primary baseline is GoLLIE~\cite{sainz2024gollie}, an LLM fine-tuned specifically to follow annotation guidelines for IE (IE) tasks. Additionally, we assess the impact of our dataset by comparing a base LLM without any manually annotated data to its counterpart trained with \guidex{}. Specifically, we evaluate Llama 3.1\footnote{We enforce constrained decoding to ensure valid outputs. While not optimal, this approach has demonstrated competitive performance for base LLMs.} against Llama 3.1 + \guidex{}, as well as GoLLIE 3.1 against GoLLIE 3.1 + \guidex{}. This allows us to isolate the contribution of our synthetic dataset on both general-purpose LLMs and task-specific fine-tuned models.

Beyond these controlled comparisons, we benchmark our approach against seven state-of-the-art models. We include general-purpose conversational LLMs such as Vicuna~\cite{vicuna2023} and ChatGPT~\cite{ouyang2022traininglanguagemodelsfollow}, as reported by \citet{zhou2023universalner}. Additionally, we evaluate against specialized IE models, including USM~\cite{10.1609/aaai.v37i11.26563}, InstructUIE~\cite{wang2023instructuiemultitaskinstructiontuning}, GoLLIE~\cite{sainz2024gollie}, KnowCoder~\cite{li2024knowcoder}, and GLiNER~\cite{zaratiana2023glinergeneralistmodelnamed}. We also compare our method to UniNER~\cite{zhou2023universalner}, a synthetic data generation-based approach for NER.

% \begin{enumerate}
%     \item Vicuna~\cite{vicuna2023} results from: ~\cite{zhou2023universalner}
%     \item USM~\cite{10.1609/aaai.v37i11.26563}
%     \item ChatGPT~\cite{ouyang2022traininglanguagemodelsfollow} results from: ~\cite{zhou2023universalner}
%     \item InstructUIE~\cite{wang2023instructuiemultitaskinstructiontuning}
%     \item UniNER ~\cite{zhou2023universalner}
%     \item KnowCoder~\cite{li2024knowcoder}
%     \item GLiNER~\cite{zaratiana2023glinergeneralistmodelnamed}
% \end{enumerate}

\subsection{Implementation details}

%Alongside GoLLIE 3.1, the Llama 3.1-based version of GoLLIE, we trained two additional models: \guidex{Llama} and \guidex{GoLLIE}. The \guidex{Llama} model consists of a Llama 3.1 instance fine-tuned exclusively on the \guidex{} dataset, while \guidex{GoLLIE} builds upon \guidex{Llama} by undergoing further fine-tuning following the GoLLIE methodology on the training datasets.

All models were trained using QLoRA~\cite{10.5555/3666122.3666563}, with hyperparameters optimized based on the validation splits of the training datasets (see Appendix \ref{sec:appendix-Implementation-Details}). We followed the same code-based input format as ~\citet{sainz2024gollie}. The data generation process was conducted on four NVIDIA A100 GPUs with 80GB of memory each, while model training was performed using two GPUs of the same type.

% \begin{itemize}
%     \item LoRA
%     \item Hiperparameters
%     \item Infrastructure 
% \end{itemize}

\begin{table*}[ht]
    \centering
    \setlength{\tabcolsep}{4pt} % Reduce column padding
    \renewcommand{\arraystretch}{1.2} % Adjust row height
    \newcommand{\cmark}{\textcolor{green}{\ding{51}}} % green checkmark
    \newcommand{\xmark}{\textcolor{red}{\ding{55}}} % red crossmark
    %\rowcolors{2}{gray!3}{white}
    \resizebox{1\textwidth}{!}{%
    \begin{tabular}{cc|ccccccc|c}
    \toprule
        \textbf{\guidex{FT}} & \textbf{Gold\textsubscript{FT}} & \textbf{AI} & \textbf{Literature} & \textbf{Music} & \textbf{Science} & \textbf{Politics} & \textbf{Movie} & \textbf{Restaurant} & \textbf{AVERAGE} \\
        \midrule
        \xmark & \xmark & 24.13 & 25.83 & 33.87 & 21.67 & 31.94 & 40.86 & 32.27 & 30.08 \\
        \cmark & \xmark 
  & 35.30\,$\pm\mathsmaller{3.2}$ 
  & 42.35\,$\pm\mathsmaller{3.6}$ 
  & 40.17\,$\pm\mathsmaller{7.73}$ 
  & 29.28\,$\pm\mathsmaller{5.3}$ 
  & 36.50\,$\pm\mathsmaller{2.0}$ 
  & 31.62\,$\pm\mathsmaller{3.9}$ 
  & 44.78\,$\pm\mathsmaller{0.9}$ 
  & 37.14\,$\pm\mathsmaller{3.4}$ \\
        \xmark & \cmark 
  & \textbf{62.56\,$\pm\mathsmaller{1.7}$}
  & \textbf{65.41\,$\pm\mathsmaller{2.0}$ }
  & 67.40\,$\pm\mathsmaller{2.77}$
  & 60.14\,$\pm\mathsmaller{0.2}$
  & 66.29\,$\pm\mathsmaller{1.9}$
  & 63.58\,$\pm\mathsmaller{0.8}$ 
  & 53.98\,$\pm\mathsmaller{1.3}$
  & 62.77\,$\pm\mathsmaller{1.2}$ \\
        \cmark & \cmark 
  & 62.41\,$\pm\mathsmaller{1.2}$
  & 63.79\,$\pm\mathsmaller{2.8}$
  & \textbf{67.92\,$\pm\mathsmaller{0.5}$}
  & \textbf{64.59\,$\pm\mathsmaller{1.1}$}
  & \textbf{69.58\,$\pm\mathsmaller{1.3}$}
  & \textbf{65.25\,$\pm\mathsmaller{0.9}$}
  & \textbf{55.50\,$\pm\mathsmaller{1.3}$}
  & \textbf{64.15\,$\pm\mathsmaller{0.7}$} \\
        \bottomrule
    \end{tabular}%
    }
    \caption{Impact of \guidex{} fine-tuning and gold fine-tuning on zero-shot NER performance with Llama 3.1 8B. \guidex{FT} denotes finetuning on \guidex{} data, while Gold\textsubscript{FT} fine-tuning on manually annotated data. ~\cmark~ shows the presence of a training stage, ~\xmark~ shows its absence. Results are reported as F1-scores on out-of-domain datasets.}
    \label{tab:first-ft-impact}
\end{table*}

\section{Results}

In this section, we discuss our experimental findings by examining \guidex{}'s impact on performance and comparing our approach against the state-of-the-art.

\label{sec:results}
\paragraph{Impact of synthetic data.}

Table~\ref{tab:first-ft-impact} summarizes our main comparisons by dividing them into two scenarios: when no manually annotated data is available, and when it is. In the first scenario, we evaluate the raw impact of our dataset, particularly analyzing the extent to which it can assist a baseline LLM in learning the task. In the second scenario, we observe how the addition of manually annotated data (and thus domain-specific knowledge) further affects performance.

The top two rows of Table~\ref{tab:first-ft-impact} show results when no manual annotations are available for fine-tuning. Without any task-specific data points, Llama 3.1 achieves an average F1 score of 30.08 across all seven datasets. While significant, it falls short of the best performing models. However, when trained with \guidex{}, the average F1 score improves significantly, with an increase of 7.06 points. Although not all seven tasks show improvement, those that do improve see notable gains. This demonstrates that our synthetic data effectively teaches the task to a baseline model and integrates domain-specific knowledge where it matters.

The third row illustrates the impact of manually annotated data (Gold\textsubscript{FT}). As anticipated, these sentence-level annotations substantially boost results. Unlike our \guidex{} dataset—largely based on documents—both the training and evaluation sets in Gold\textsubscript{FT} focus on sentence-level IE tasks. This alignment phase allows the model to tackle sentence-specific zero-shot tasks more accurately. Even so, the improvements we achieve with Gold\textsubscript{FT} are complementary to those achieved with \guidex{FT}.

Lastly, the bottom row presents a model fine-tuned first with \guidex{FT} and then with Gold\textsubscript{FT}. 
It can be seen that both steps jointly increase performance by an average of 34 F1 points over the plain Llama 3.1 baseline (27 points over \guidex{FT} alone and 1.4 over Gold\textsubscript{FT}). Moreover, applying \guidex{FT} to a model already trained with Gold\textsubscript{FT} improves on five out of the seven datasets. This indicates that the \guidex{FT} data can significantly enhance the model's performance in various specific domains. 
We analyze the impact on a label-by-label basis in Section~\ref{sec:analysis}.

\begin{table*}[h]
\centering
\resizebox{\textwidth}{!}{%
\begin{tabular}{r|rr|ccccccc|c}

\toprule
\textbf{Model} & \textbf{Params} & \textbf{Backbone} & \textbf{Movie} & \textbf{Restaurant} & \textbf{AI} & \textbf{Literature} & \textbf{Music} & \textbf{Politics} & \textbf{Science} & \textbf{Avg} \\
\midrule
Vicuna-7B & 7B & Llama & 06.0 & 05.3 & 12.8 & 16.1 & 17.0 & 20.5 & 13.0 & 13.0 \\
Vicuna-13B & 13B & Llama & 00.9 & 00.4 & 22.7 & 22.7 & 26.6 & 27.0 & 22.0 & 17.5 \\
USM & 0.3B & & 37.7 & 17.7 & 28.2 & 56.0 & 44.9 & 36.1 & 44.0 & 37.8 \\
ChatGPT & $-$ & $-$ & 05.3 & 32.8 & 52.4 & 39.8 & 66.6 & 68.5 & \textbf{67.0} & 47.5 \\
InstructUIE & 11B & FlanT5 & 63.0 & 21.0 & 49.0 & 47.2 & 53.2 & 48.1 & 49.2 & 47.2 \\
UniNER-7B & 7B & Llama & 42.4 & 31.7 & 53.6 & 59.3 & 67.0 & 60.9 & 61.1 & 53.7 \\
UniNER-13B & 13B & Llama & 48.7 & 36.2 & 54.2 & 60.9 & 64.5 & 61.4 & 63.5 & 55.6 \\
GoLLIE & 7B & CodeLlama & 63.0 & 43.4 & 59.1 & 62.7 & 67.8 & 57.2 & 55.5 & 58.0 \\
KnowCoder & 7B & Llama 2  & 50.0 & 48.2 & 60.3 & 61.1 & \textbf{70.0} & 72.2 & 59.1 & 60.1 \\
GLiNER-L & 0.3B & DeBERTa-V3 & 57.2 & 42.9 & 57.2 & 64.4 & 69.6 & \textbf{72.6} & 62.6 & 60.9 \\
\midrule
Gold\textsubscript{FT}& 8B & Llama 3.1 & 63.6\,$\pm\mathsmaller{0.8}$ %mitmovie
  & 54.0\,$\pm\mathsmaller{1.3}$ %mitrestaurant
  & \textbf{62.6\,$\pm\mathsmaller{1.7}$} %ai
  & \textbf{65.4\,$\pm\mathsmaller{2.0}$ } %literature
  & 67.4\,$\pm\mathsmaller{2.8}$ %music
  & 66.3\,$\pm\mathsmaller{1.9}$ %politics
  & 60.1\,$\pm\mathsmaller{0.2}$  %science
  & 62.8\,$\pm\mathsmaller{1.2}$ \\ %avg 
\rowcolor{gray!10} \guidex{FT} + Gold\textsubscript{FT} & 8B & Llama 3.1 & \textbf{65.3\,$\pm\mathsmaller{0.9}$} %mitmovie
  & \textbf{55.5\,$\pm\mathsmaller{1.3}$} %mitrestaurant
  & 62.4\,$\pm\mathsmaller{1.2}$ %ai
  & 63.8\,$\pm\mathsmaller{2.8}$  %literature
  & 67.9\,$\pm\mathsmaller{0.5}$ %music
  & 69.6\,$\pm\mathsmaller{1.3}$ %politics
  & 64.6\,$\pm\mathsmaller{1.1}$ %science
  & \textbf{64.2\,$\pm\mathsmaller{0.7}$}  \\ %avg

\bottomrule
\end{tabular}
}
\caption{Zero-Shot F1-scores on Out-of-Domain NER Benchmarks, reporting state-of-the-art systems and two systems fine-tuned with and without \guidex{}. Results are averaged across 3 runs.}
\label{tab:ner_sota_benchmark}
\end{table*}

\paragraph{Comparison with the state-of-the-art.}

Table~\ref{tab:ner_sota_benchmark} presents a comparison of our best-performing model, which was fine-tuned on \guidex{FT} and Gold\textsubscript{FT}, against various state-of-the-art zero-shot NER systems.
%, including general-purpose LLMs (Vicuna~\cite{vicuna2023}, ChatGPT~\cite{ouyang2022traininglanguagemodelsfollow}), instruction-tuned models (InstructUIE~\cite{wang2023instructuiemultitaskinstructiontuning}, UniNER~\cite{zhou2023universalner}), and IE-specific models (GoLLIE~\cite{sainz2024gollie}, KnowCoder~\cite{li2024knowcoder}, GLiNER~\cite{zaratiana2023glinergeneralistmodelnamed}).
Our best approach achieves an average F1 score of 64.2, surpassing all other models in the benchmark. Notably, it outperforms GoLLIE by 6.2 F1 points. GoLLIE is a system similar to Gold\textsubscript{FT} but based on CodeLlama. Additionally, when compared to KnowCoder, another system akin to Gold\textsubscript{FT} that uses a pretraining dataset to better follow annotation schemas, our approach shows a 4.1-point improvement. It is worth mentioning that the pretraining proposed by ~\citet{li2024knowcoder} could provide complementary enhancements to our method.
In addition to their overall performance, models trained on \guidex{} show strong generalization across various domains, achieving the highest F1 scores in two out of seven benchmarks: Movie (65.3), Restaurant (55.5), and being the best overall. Particularly in Politics, our model achieves a +12.4 F1 point improvement over GoLLIE (57.2), showcasing its ability to capture domain-specific nuances. Even in Music, where GLiNER-L slightly outperforms our model (69.6 vs. 67.9 F1), \guidex{} remains competitive, despite GLiNER's explicit focus on generalist NER modeling.

These results highlight the effectiveness of our approach in adapting to unfamiliar domains, surpassing conventional instruction tuning and guideline-based baselines. By utilizing LLMs and domain-specific documents to generate synthetic data, \guidex{} offers a reliable method for enhancing zero-shot IE.

\section{Analysis} \label{sec:analysis}

\begin{table*}[!htb]
\centering
\vspace{-1.0em}
\small
\renewcommand{\arraystretch}{1.5} % Increase vertical spacing between rows
    \resizebox{\textwidth}{!}{%
    \begin{tabular}{llp{7.5cm}cc}
        \multicolumn{4}{c}{} \\
        \toprule
        \multicolumn{1}{l}{\textbf{Dataset}} & \multicolumn{1}{l}{\textbf{Label}} & \multicolumn{1}{l}{\textbf{Summarized Guideline}} & \textbf{Gold\textsubscript{FT}} & \textbf{ \guidex{} + Gold\textsubscript{FT}} \\ \midrule
        \rowcolor{ForestGreen!10} Natural Science & Scientist & A person who is studying or has expert knowledge of a natural science field. & 38.43 & 51.21 \\
        \rowcolor{ForestGreen!10}  & Person & Individuals that are not scientist. & 48.53 & 53.02 \\ \hline
        \rowcolor{ForestGreen!10} Politics & Politician & A person who is actively engaged in politics, holding a public office, involved in political activities or part of a political party.  & 35.12 & 44.37 \\
        \rowcolor{ForestGreen!10}  & Person & Individuals that are not politician. & 59.48 & 62.86 \\\hline
        \rowcolor{ForestGreen!10} Politics & PoliticalParty & An organization that coordinates candidates to compete in a particular country's elections.  & 58.55 & 65.30 \\
        \rowcolor{ForestGreen!10}  & Organization & A structured group, institution, company, or association that is not a political party. & 54.34 & 58.38 \\ \hline
        \rowcolor{CornflowerBlue!10} AI & Location & A specific geographical or structural location. & 75.36 & 75.86 \\
        \rowcolor{CornflowerBlue!10} Music & Country & A sovereign nation. & 81.62 & 82.45 \\ \hline
        % \rowcolor{Thistle!10} Literature & LiteraryGenre & Categories in literature defined by unique artistic techniques, themes, content, and lengths. & 46.73 & 42.73 \\
        % \rowcolor{Thistle!10} MITRestaurant & Rating & Any specific qualitative or quantitative evaluation of a restaurant's quality, performance, or popularity. & 34.80 & 31.38 \\
        \rowcolor{Thistle!10} Music & Other & Named entities that are not included in any other category.  & 15.99 & 13.02 \\
        \rowcolor{Thistle!10} Literature & Other & Named entities that are not included in any other category.  & 23.19 & 20.78 \\
        \bottomrule
    \end{tabular}
    }
    %\caption{F1 scores for specific labels from different datasets with summarized guideline descriptions. On \colorbox{ForestGreen!20}{green} the labels where the domain knowledge acquired by training in \guidex{} is helpful for the model. On \colorbox{CornflowerBlue!20}{blue} the labels where there is no improvement. And, on \colorbox{Thistle!20}{red}, those labels that both systems struggle to identify correctly.}
    \caption{F1 scores for specific labels from different datasets with summarized guideline descriptions. On green, the labels where the domain knowledge acquired by training in \guidex{} is helpful for the model. On blue, the labels where there is no improvement. And, on red, those labels that both systems struggle to identify correctly.}
    \label{tab:error_analysis_summarized}
\end{table*}

This section examines the impact of \guidex{} across different labels, highlighting its benefits and limitations. Table~\ref{tab:error_analysis_summarized} provides a breakdown of performance gains in various domains and identifies cases where \guidex{} still struggles.

% \paragraph{When does domain data help?} Previous studies have shown that, in new domains, models often assign known labels (those present in the training data) over new, domain-specific labels~\cite{sainz2024gollie}.Baseline models frequently default to generic labels like \textit{Person} instead of recognizing domain-specific entities such as \textit{Scientist} and \textit{Politician}, leading to misclassifications. Fine-tuning with \guidex{} significantly improves precision, with F1 gains of up to 12.8 points for these cases, as well as a 6.75-point increase in distinguishing \textit{PoliticalParty} from the broader \textit{Organization} category. The model, now equipped with explicit contextual definitions, applies labels more accurately, reducing errors where political parties were misclassified as generic organizations due to linguistic similarities. These results highlight the impact of structured guideline-driven learning in improving model adaptability, reinforcing context-aware predictions, and enabling more precise entity differentiation in specialized domains.

\paragraph{Do Guidelines Improve Domain-Specific Labels?} \guidex{} effectively mitigates the overgeneralization tendency in zero-shot IE~\cite{sainz2024gollie} by teaching models to differentiate between broad and fine-grained entity labels through structured annotation schemas. Table~\ref{tab:error_analysis_summarized} shows how baseline models frequently default to generic labels like \textit{Person} instead of recognizing domain-specific entities such as \textit{Scientist} and \textit{Politician}, leading to misclassifications. Fine-tuning with \guidex{} significantly improves precision, with F1 gains of up to 12.8 points for these cases, as well as a 6.75-point increase in distinguishing \textit{PoliticalParty} from the broader \textit{Organization} category. The model, trained on explicit contextual definitions, applies labels more accurately, reducing errors where, for instance, political parties were misclassified as organizations due to linguistic similarities. These results highlight the impact of structured guideline-driven learning in improving model adaptability, reinforcing context-aware predictions, and enabling more precise entity differentiation in specialized domains. For inherently generic labels, such as \textit{Location} and \textit{Country}, the model already achieves strong performance without \guidex{}. This suggests that our approach is most beneficial for refining entity granularity rather than improving well-established, domain-agnostic categories.

% \paragraph{Remaining challenges.} As shown in Figure~\ref{tab:error_analysis_summarized}, there are distinct cases where neither the guidelines nor the \guidex{} data are sufficient for accurately identifying labels, such as \textit{Other} or \textit{Miscellaneous}. These labels lack clear definitions, which renders the associated guidelines less effective. This challenge is intensified when training the model using the specific guidelines outlined in \guidex{}, as they are intentionally designed to be very precise. However, this issue is not new; it has been previously recognized in models trained to follow guidelines~\cite{sainz2024gollie}.

\paragraph{Remaining challenges.} Some labels, such as \textit{Other} and \textit{Miscellaneous}, remain problematic even with \guidex{} (see Table~\ref{tab:error_analysis_summarized}). These categories often lack clear definitions, making it difficult to apply them consistently. Since \guidex{} generates precise guidelines, the absence of well-defined annotation criteria for these broad labels limits its effectiveness. This aligns with findings~\cite{sainz2024gollie} suggesting that guideline-driven models struggle with vague or catch-all entity types.

\section{Conclusions}

In this paper, we introduce \guidex{}, a novel approach for synthetic data generation aimed at IE. We utilize \guidex{} to generate data suitable for a variety of domains using documents from FineWeb-edu. As a demonstration of the method, we generate a dataset with 10,000 annotated documents, featuring a wide range of labels, from generic to highly domain-specific. Using this generated data, we train an IE model that surpasses the performance of current state-of-the-art zero-shot NER systems. Furthermore, we demonstrate that our method effectively generates domain-specific annotations, which can be utilized to train robust IE systems across multiple domains.

% Our method opens the way for further advances, including document-level IE method creation, as well as the development of automatic data generation techniques for poorly defined or generic labels, such as \textit{Other} or \textit{Miscellaneous}. \guidex{} offers a clear pathway to tackle both aforementioned areas of research, which remain largely unexplored and present a significant opportunity.

\guidex{} paves the way for two key research directions: (1) advancing document-level IE methodologies, and (2) developing automated techniques for handling ill-defined or generic labels like \textit{Other} and \textit{Miscellaneous}. These challenges are critical for IE applications but remain largely unexplored, offering potential for future work. Our method provides a foundation to address these open problems systematically.

Finally, it is worth noting that all four prompts, the driving script, and the consistency filter act as drop-in tools that could be pointed at any plain-text corpus. GuideX will return a ready-to-train dataset within hours.

\section*{Limitations}

In this work, we present an approach for generating synthetic data, which we used to enhance the performance of IE models on NER datasets across various domains. However, our evaluation does not encompass all potential applications of this approach. For example, the automatically generated dataset consists of document-level texts rather than individual sentences. Our evaluation framework, on the other hand, focuses solely on sentence-level tasks. In the future, we aim to investigate the impact of our approach on document-level tasks. 

A second limitation is the use of catch-all tags such as \textit{Other} and \textit{Miscellaneous} highlighted in Table~\ref{tab:error_analysis_summarized}. These categories function as residual classes, aggregating semantically diverse spans that lack a consistent boundary. As a result, they introduce ambiguity and noise during training. A possible direction for future work is to encode these spans into a shared embedding space, apply unsupervised clustering, and use an LLM to induce candidate schema definitions for coherent subsets. This would enable a second GuideX iteration, replacing vague categories with more fine-grained, guideline-compatible types.

\section*{Acknowledgements}
This work has been partially supported by the Basque Government (Research group funding IT1570-22 and IKER-GAITU project). We are also thankful to several projects funded by MCIN/AEI/10.13039/501100011033: (i) DeepKnowledge (PID2021-127777OB-C21) and by FEDER, EU; and AWARE (TED2021-131617B-I00) and by the European Union NextGenerationEU/PRTR. Neil De La Fuente was supported by the Basque Government through the IKASIKER 2025 scholarship programme.

% Bibliography entries for the entire Anthology, followed by custom entries
%\bibliography{anthology,custom}
% Custom bibliography entries only
\bibliography{custom}

\clearpage
\newpage
\appendix

\section{Implementation Details}
\label{sec:appendix-Implementation-Details}
In this section, we further detail the whole process of \guidex{}. In Section~\ref{sec:appendix-guidex_dataset_gen} we explain how we built the dataset, focusing on the hyperparameters used, the multi-step generation prompts, and the filtering process that ensures consistent annotations. In \ref{sec:appendix_guidexft_goldft} we describe the hyperparameters for both \guidex{FT} and Gold\textsubscript{FT}, outlining how they were tuned to achieve the outcomes showcased on Section~\ref{sec:results}.

\subsection{\guidex{} Dataset generation}
\label{sec:appendix-guidex_dataset_gen}

The \guidex{} dataset was built through a structured multi-step process designed to ensure high-quality, consistent annotations and guidelines. The dataset generation pipeline follows a systematic approach involving prompt-based multi-step generation, filtering for consistency, and the use of hyperparameter tuning to optimize the outputs.

\paragraph{Model and Hyperparameters.} The synthetic dataset was generated using Llama 3.1-70B Instruct, leveraging vLLM for efficient inference. The detailed hyperparameter settings used in the generation process are available in Table 5. 

\begin{figure*}[!ht]
    \centering
    \includegraphics[width=1\linewidth]{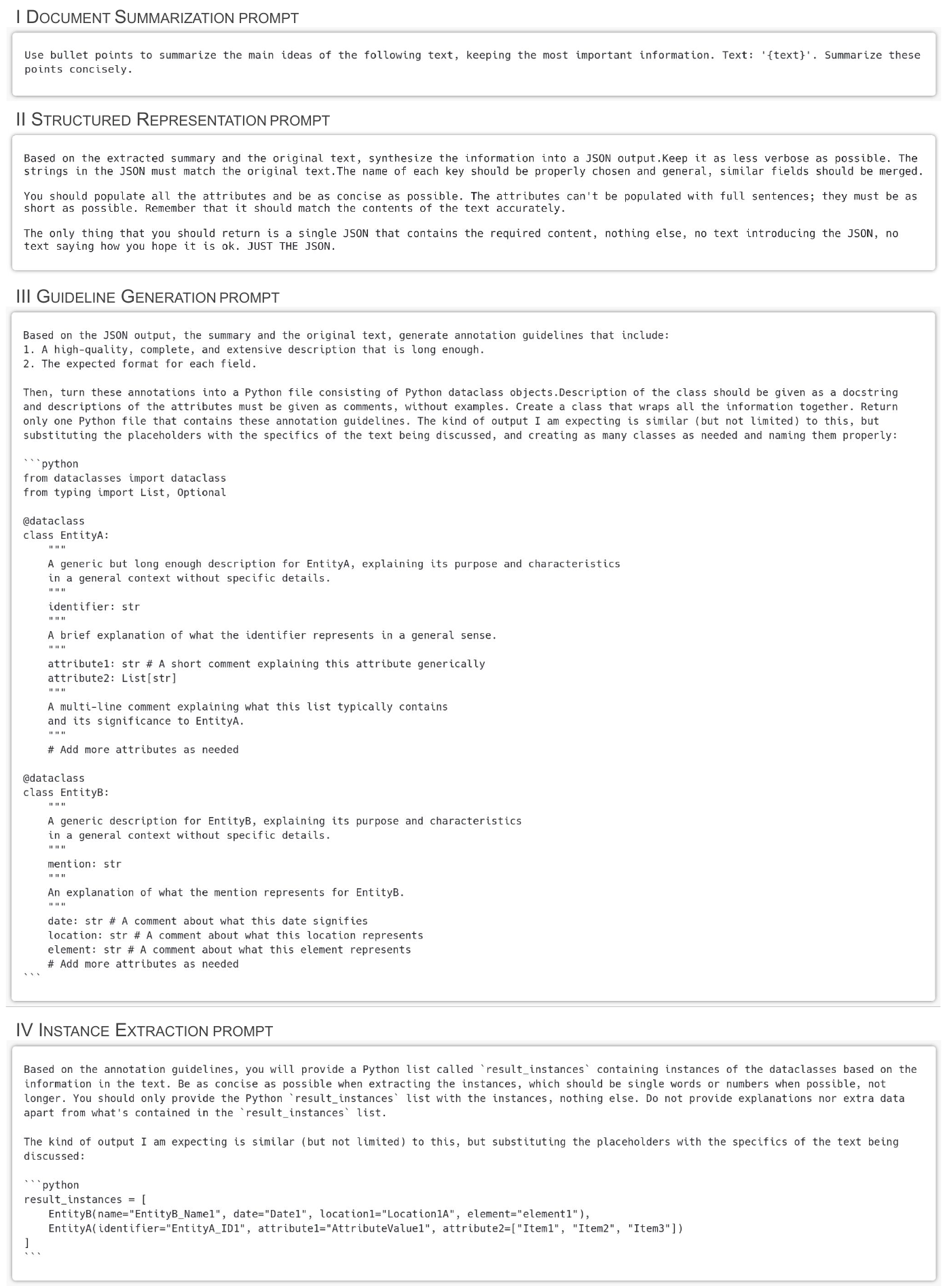}
    \caption{\guidex{} follows a multi-step prompting pipeline which allows for the creation of the synthetic guidelines and annotations that conform \guidex{} and are used for \guidex{FT}.}
    \label{fig:synthetic-data-prompts}
\end{figure*}

\paragraph{Multi-Step Generation Process.} To ensure structured and meaningful outputs, the dataset was built using a four-step generation pipeline. The first step involved extracting key points from the input text, reducing redundancy while preserving essential information. The second step transformed this summarized content into a structured JSON representation, ensuring a consistent and standardized format. Next, annotation guidelines were generated to define the expected attributes and structure, facilitating consistency across all annotations. Finally, the model extracted instances based on these guidelines, ensuring that the final dataset adhered to a coherent format. The complete set of prompts used for each step can be found in Figure~\ref{fig:synthetic-data-prompts}.

\begin{table}[!t]

\centering
\small
\setlength{\tabcolsep}{6pt}  
\renewcommand{\arraystretch}{1.1}  

\begin{adjustbox}{width=0.49\textwidth}
\begin{tabular}{l|l|l}
\toprule
\textbf{Category} & \textbf{Hyperparameter} & \textbf{Value} \\
\midrule
\multirow{5}{*}{\textbf{Model Setup}}
 & Model Name & Llama 3.1-70B Instruct \\
 & Tokenizer & Llama 3.1-70B Instruct \\
 & Dtype & \texttt{bfloat16} \\
 & Max Seq. Length & 8192 tokens \\
 & Tensor Parallel Size & 2 \\
\midrule
\multirow{3}{*}{\textbf{Generation Config}}
 & Temperature & 0.7 \\
 & Top-p & 0.95 \\
 & Max New Tokens & 1024 \\
\midrule
\multirow{2}{*}{\textbf{Batching}}
 & Batch Size & 32 \\
 & Processing Mode & Batched prompts via \texttt{vLLM} \\
\midrule
\multirow{3}{*}{\textbf{Hardware}}
 & GPUs Used & 4× A100 (80GB) \\
 & CPU per Task & 16 \\

\bottomrule
\end{tabular}
\end{adjustbox}
\caption{Hyperparameter Settings for the \guidex{} Synthetic Data Generation Stage.}
\label{tab:guidex_generation_hparams}
\end{table}

% \section{Dataset Details and Overlap Analysis}
% \label{sec:appendix_data_overlap}
% Table~\ref{tab:dataset-overlap} details the datasets used for training and evaluation, along with statistics on the overlap of entity types between the gold datasets and our synthetically generated \guidex{} dataset. The overlap percentage indicates the proportion of gold entity types that are also present (by name) in the \guidex{} dataset schema. Note that the 'Train Instances' column reflects the number of processed lines from the JSON analysis file, as sentence counts were not readily available for all splits. Dev/Test instance counts are derived similarly where available.

%TABLE TO ADD 

\subsection{\guidex{FT} \& Gold\textsubscript{FT}}
\label{sec:appendix_guidexft_goldft}

This section summarizes the fine-tuning configurations for \guidex{FT} and Gold\textsubscript{FT}. Both models use Llama 3.1-8B with 4-bit LoRA, AdamW optimization, and a cosine scheduler. \guidex{FT} supports longer sequences (8192 tokens) as it is suited for document-level input texts, and employs gradient accumulation, while Gold\textsubscript{FT} uses a larger per-device batch size without accumulation. Training is conducted on 2× A100 (80GB) GPUs with DeepSpeed Zero-3, differing in sequence length, batch sizes, and the number of training epochs. A full overview is provided in Table~\ref{tab:hyperparams}.

\begin{table*}[t]
\centering
\small
\setlength{\tabcolsep}{6pt} 
\renewcommand{\arraystretch}{1.}  
\begin{adjustbox}{width=\textwidth}
\begin{tabular}{l|l|c|c}
\toprule
\textbf{Category} & \textbf{Hyperparameter} & \textbf{\guidex{FT}} & \textbf{Gold\textsubscript{FT}} \\
\midrule
\multirow{5}{*}{\textbf{Model \& Quantization}} 
& Base Model & Llama 3.1-8B & Llama 3.1-8B \\
& Quantization & 4-bit LoRA & 4-bit LoRA \\
& LoRA Rank ($r$) & 128 & 128 \\
& LoRA $\alpha$ & 256 & 256 \\
& LoRA Dropout & 0.08 & 0.05 \\
& Dtype & \texttt{bfloat16} & \texttt{bfloat16}  \\
\midrule
\multirow{5}{*}{\textbf{Optimization}} 
& Optimizer & AdamW & AdamW \\
& Learning Rate & $3 \times 10^{-4}$ & $3 \times 10^{-4}$ \\
& Weight Decay & 0.001 & 0.001 \\
& Scheduler & Cosine & Cosine \\
& Warmup Steps & 10\% of total steps & 10\% of total steps \\
\midrule
\multirow{3}{*}{\textbf{Batching  \& Seq. Length}} 
& Per-device Batch Size & 4 & 16 \\
& Gradient Accumulation Steps & 2 & None \\
& Effective Batch Size & 8 & 16\\ 
& Max Sequence Length (tokens) & 8192 & 2048 \\
\midrule
\multirow{2}{*}{\textbf{Epochs \& Checkpoints}} 
& Epochs & 3 & 1 \\
& Checkpoint Strategy & End of each epoch & End of each epoch \\
\midrule
\multirow{2}{*}{\textbf{Hardware}} 
& GPUs Used & 2× A100 (80GB) & 2× A100 (80GB) \\
& Multi-GPU Support & DeepSpeed Zero-3 & DeepSpeed Zero-3 \\
& CPU per Task & 22 & 22 \\
\bottomrule
\end{tabular}
\end{adjustbox}

\caption{Hyperparameter Settings for \guidex{FT} and Gold\textsubscript{FT}.}
\label{tab:hyperparams}
\end{table*}

\end{document}